\documentclass[11pt,a4paper]{article}
\usepackage[hyperref]{eacl2021}
\usepackage{times}
\usepackage{latexsym}

\usepackage{subcaption}
\usepackage{caption}
\usepackage{graphicx}
\usepackage{natbib}
\usepackage{amsmath}
\usepackage{xcolor}
\usepackage{subcaption}

\usepackage{cleveref}
\usepackage{microtype}

\aclfinalcopy 
\def\aclpaperid{855} 

\usepackage{amsthm}
\DeclareUnicodeCharacter{2212}{-}

\swapnumbers 
\newtheorem{thm}{Theorem}[section] 

\theoremstyle{plain} 
\newcommand{\thistheoremname}{}
\newtheorem{genericthm}[thm]{\thistheoremname}

\title{Paraphrases do not explain word analogies}

\author{Louis Fournier \\
  Cognitive Machine Learning \\
  CNRS-EHESS-INRIA\\
  PSL Research University \\
  Paris, France \\
  \texttt{louis.fournier@polytechnique.edu} \\\And
  Ewan Dunbar \\
  Cognitive Machine Learning \\
  CNRS-EHESS-INRIA\\
  PSL Research University \\
  Paris, France \\
  and \\
  University of Toronto \\
  Toronto, Canada\\
  \texttt{ewan.dunbar@utoronto.ca} \\}

\date{}

\begin{document}
\maketitle
\begin{abstract}
Many types of distributional word embeddings (weakly) encode linguistic regularities as directions (the difference between \emph{jump} and \emph{jumped} will be in a similar direction to that of \emph{walk} and \emph{walked}, and so on). Several attempts have been made to explain this fact. We respond to Allen and Hospedales' recent (ICML, 2019) theoretical explanation, which claims that word2vec and GloVe will encode linguistic regularities  whenever a specific relation of \emph{paraphrase} holds between the four words involved in the regularity. We demonstrate that the explanation does not go through: 
the paraphrase relations needed under this explanation do not hold empirically.   

\end{abstract}

\section{Introduction}

The study of linguistic regularities in distributional word embeddings---that the difference vector calculated between the vectors \emph{jump} and \emph{jumped} shows a similar direction to that of \emph{walk} and \emph{walked}, and so on---has been both stimulating and controversial. While a number of such regularities appear to hold, across a number of different kinds of embeddings, the standard  \textsc{3CosAdd} analogy test used to measure the presence of these regularities has come under fire  for confounding analogical regularities with unrelated properties of semantic embeddings. It is thus important to note that several papers have proposed theoretical explanations for why linguistic regularities \emph{should} hold in distributional word embeddings. Particularly in light of the controversies over linguistic regularities, it is important to examine the soundness of these arguments.

\citet{allen} develop such an explanation by linking the semantic definition of an analogy to \emph{paraphrases.} In the sense of \citet{gittens-etal-2017-skip}, paraphrases are sets of words which are semantically and distributionally closely equivalent to another word or set of words---for example, \emph{king} may be paraphrased by \{\emph{man}, \emph{royal}\}. Allen and Hospedales argue that the standard analogy criterion, that \emph{king} - \emph{man} + \emph{woman} = \emph{queen},  is equivalent to a criterion whereby \{\emph{king}, \emph{woman}\} paraphrases \{\emph{man}, \emph{queen}\}. With this in mind, it becomes possible to rewrite the arithmetic analogy criterion in terms of vectors encoding the pointwise mutual information (PMI) between words and their contexts, and to decompose the error in the analogy equality into several components, including a \emph{paraphrase error} term  measuring the degree to which the critical paraphrase holds. Making use of an assumption that the word2vec embedding is a linear transformation of the PMI matrix, they argue that  results in terms of PMI apply to word vectors. Thus, under their explanation, a major part of  success  on an analogy $a-a^*+b^*=b$ is due to ${a,b^*}$ and ${a^*,b}$ being close distributional paraphrases.
 

We first review the literature on the analogy test itself, underlining known pitfalls which any  explanation of linguistic regularities must navigate.
We then show empirically that the relation between the PMI matrix and word2vec embeddings is
to some degree
linear,
which may be enough to satisfy
the assumption of \citet{allen}.
We further examine the proposed decomposition into error terms. We demonstrate that, empirically, these error terms tend to be undefined due to data sparseness, undermining their explanatory force. Most importantly, examining a number of analogies which pass the standard test, we show that the critical paraphrase error term is, contrary to the proposed explanation, very large.\footnote{Code is available at \url{www.github.com/bootphon/paraphrases_do_not_explain_analogies}.}



\section{Related work}

 Early works proposing explanations of the analogical properties of word embeddings include \citet{MikolovYihEtAl_2013_Linguistic_Regularities_in_Continuous_Space_Word_Representations} and \citet{pennington-etal-2014-glove}. A geometrical explanation is proposed by \citet{arora-etal-2016-latent}, but this explanation relies on very strong preconditions, notably, that the word vectors be distributed uniformly in space. 
 \citet{ethayarajh-etal-2019-towards} also propose an explanation, providing a link between the PMI and the norm of word embeddings. However, as pointed out by \citet{allen}, this explanation, too, rests on strong assumptions. Notably, the words involved in the analogy are required to be coplanar, a property that seems unlikely in light of the lack of parallelism we discuss in the next section. 

\section{Issues with the test}


Issues have arisen with the  standard way of measuring linguistic analogies. 
\citet{levy-goldberg-2014-linguistic}, \citet{vylomova-etal-2016-take},  \citet{RogersDrozdEtAl_2017_Too_Many_Problems_of_Analogical_Reasoning_with_Word_Vectors}, and \citet{fournier-etal-2020-analogies} 
all demonstrate that the standard \textsc{3CosAdd} measure conflates  several very different properties of  embeddings, simultaneously measuring not only the directional regularities suggested by typical illustrations of vectors   in a parallelogram, but also the similarity of individual matched pairs such as \emph{king, man}, as well as the global arrangement of vectors in  semantic fields, such as  \emph{king, queen, prince, \dots} versus  \emph{man, woman, child, \dots} in distinct regions of the space. These issues undermine the construct validity of the standard analogy test.  This conflation of  properties explains certain pathological behaviours of the test  \cite{Linzen_2016_Issues_in_evaluating_semantic_spaces_using_word_analogies,RogersDrozdEtAl_2017_Too_Many_Problems_of_Analogical_Reasoning_with_Word_Vectors}. In spite of these issues, \citet{fournier-etal-2020-analogies} demonstrate, using alternative measures, that linguistic regularities are nevertheless coded by directional similarities. This parallelism is weak, with directions tending to be closer, in the absolute, to being orthogonal than to being parallel, but is present above chance level  (unmatched word pairs).

Thus, before turning to \citet{allen}, one of a number of theoretical attempts to explain performance on the \textsc{3CosAdd} objective, we underscore that such demonstrations run the risk of explaining properties of the test which may be of secondary interest, or, conversely, of placing undue emphasis on the role of directional regularities, which have been shown to play only a small role in success on  \textsc{3CosAdd}.

\section{Explaining analogies through paraphrases}


For a word $w_i$ and a word $c_j$ which can appear in the context of $w_i$, the pairwise mutual information $PMI(w_i,c_j)$ is defined as $\log\frac{p(w_i,c_j)}{p(w_i)p(c_j)}$. As shown by \citet{w2v_pmi_goldberg}, skip-gram word2vec with negative sampling factorizes the PMI: PMI $\approx W^\top\cdot C$, with $W$ and $C$ the word and context embedding matrices of a word2vec model.

For  two pairs of words $(a,a^*)$ and $(b,b^*)$ from the same semantic relation, the standard arithmetic analogy test criterion is that $a - a^* +b^* = b$. Writing $\mathcal{W}=\{a,b^*\}, \mathcal{W_*}=\{a^*,b\}$, and $\textbf{PMI}_{x}$ the PMI vector of $x$,   \citet{allen} show that is possible to rewrite the arithmetic analogy formula with PMI vectors, and to decompose the error in the equality into five terms as follows:

\begin{equation}
\begin{split}
\textbf{PMI}_{b^*} = & \textbf{PMI}_b + \textbf{PMI}_{a^*} - \textbf{PMI}_a \\
+ & \underbrace{\rho^{\mathcal{W, W_*}}}_{\text{Paraphrase error}} + \underbrace{\sigma^\mathcal{W} - \sigma^{\mathcal{W_*}}}_{\text{Conditional dependence error}} \\
+ & \underbrace{(\tau^\mathcal{W} - \tau^{\mathcal{W}_*})\textbf{1}}_{\text{Mutual dependence error}}
\end{split}
\label{eq_allen}
\end{equation}
\normalsize

 The error terms are vectors of length $\left|\mathcal{V}\right|$ (vocabulary size), with each element $j$ defined as:

\begin{equation}
\begin{split}
\rho^{\mathcal{W, W_*}} &= \log\frac{p(c_j|\mathcal{W_*})}{p(c_j|\mathcal{W})} \\
\sigma^\mathcal{W} &= \log\frac{p(\mathcal{W}|c_j)}{\prod_\mathcal{W} p(w_i|c_j)} \\
\tau^\mathcal{W} &= \log\frac{p(\mathcal{W})}{\prod_\mathcal{W} p(w_i)}
\end{split}
\end{equation}

\normalsize
The authors claim that these terms can be embedded linearly into a word2vec embedding space by multiplying them by the Moore-Penrose pseudo-inverse $C^\dag$ of the context matrix $C$. Then with $\textbf{w}_x$ the word2vec embedding of $x$, $C^{\dag} \cdot \textbf{PMI}_{x} \approx \textbf{w}_x$. Thus we get the final decomposition:

\begin{equation}
\begin{split}
\textbf{w}_{b^*} = & \textbf{w}_b + \textbf{w}_{a^*} - \textbf{w}_a + \\ 
& C^{\dag} \left( \rho^\mathcal{W, W_*} + \sigma^\mathcal{W} - \sigma^\mathcal{W_*} - (\tau^\mathcal{W} - \tau^\mathcal{W_*})\textbf{1} \right)
\end{split}
\end{equation}

\normalsize
The paraphrase error term  is claimed to be small for successful analogies. Elaborating on the notation, $\mathcal{W}$ is taken to paraphrase $\mathcal{W_*}$ if, wherever all  $w\in\mathcal{W}$ appear together, we observe the same distribution of surrounding words as for  $\mathcal{W_*}$. The paraphrase error assesses the similarity of the  distributions of words in the context of $\mathcal{W}$ (all words in $\mathcal{W}$ appearing together) versus $\mathcal{W_*}$.


\section{Linearity of the link between PMI and word2vec \label{section pmi}}


Though it is true that there is a relation between the  word2vec matrices $W^\top\!\cdot\!C$ and the PMI matrix, in practice the link is more complicated than simple linear matrix factorization, due in part to the training tricks described in \citet{MikolovChenEtAl_2013_Efficient_estimation_of_word_representations_in_vector_space}.  
 The result of \citet{allen} requires that the 
 embedding from PMI vectors to word2vec embeddings be ``linear enough'' for $C^\dag\cdot$ PMI to approximate $W$.
 
\begin{figure}
\centering
\captionsetup[subfigure]{labelformat=empty}
\begin{subfigure}[t]{0.48\textwidth}
\centering
\includegraphics[width=\textwidth]{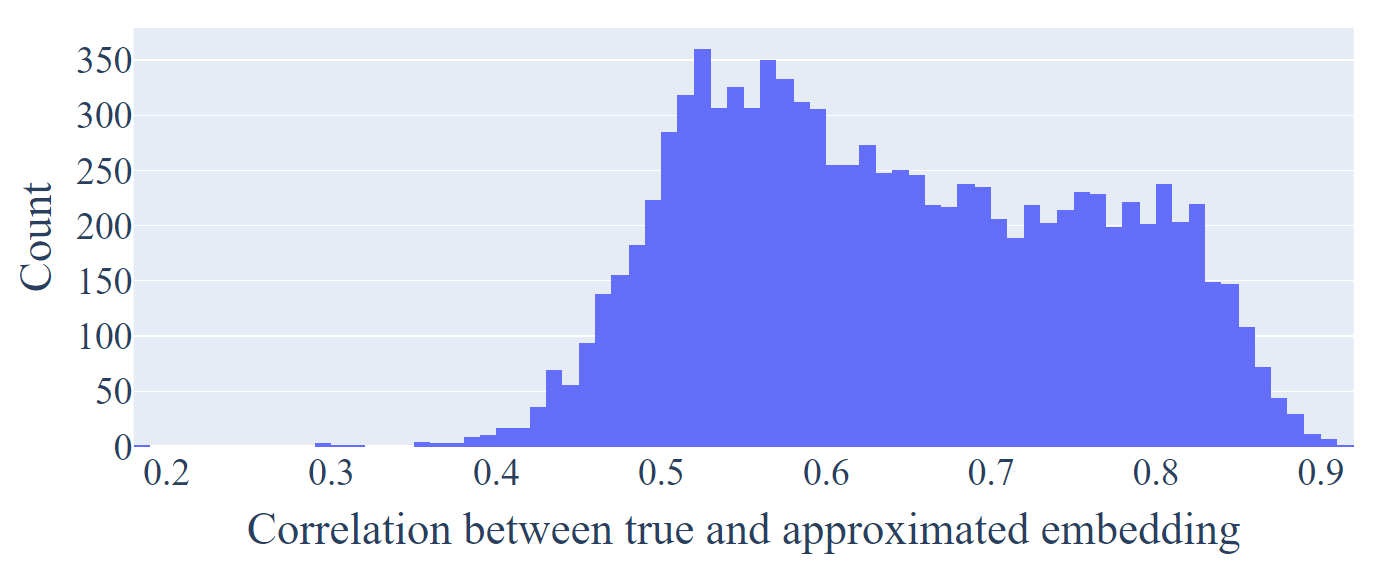}
\caption{Figure 1a: Histogram of the Pearson correlations between true and approximated word2vec embeddings for the top ten thousand words in the \emph{text8} corpus. The mean value is $0.643$ and the variance is $0.014$.}
\label{plot link spearman}
\end{subfigure}
\begin{subfigure}{0.48\textwidth}
\centering
\includegraphics[width=.8\textwidth]{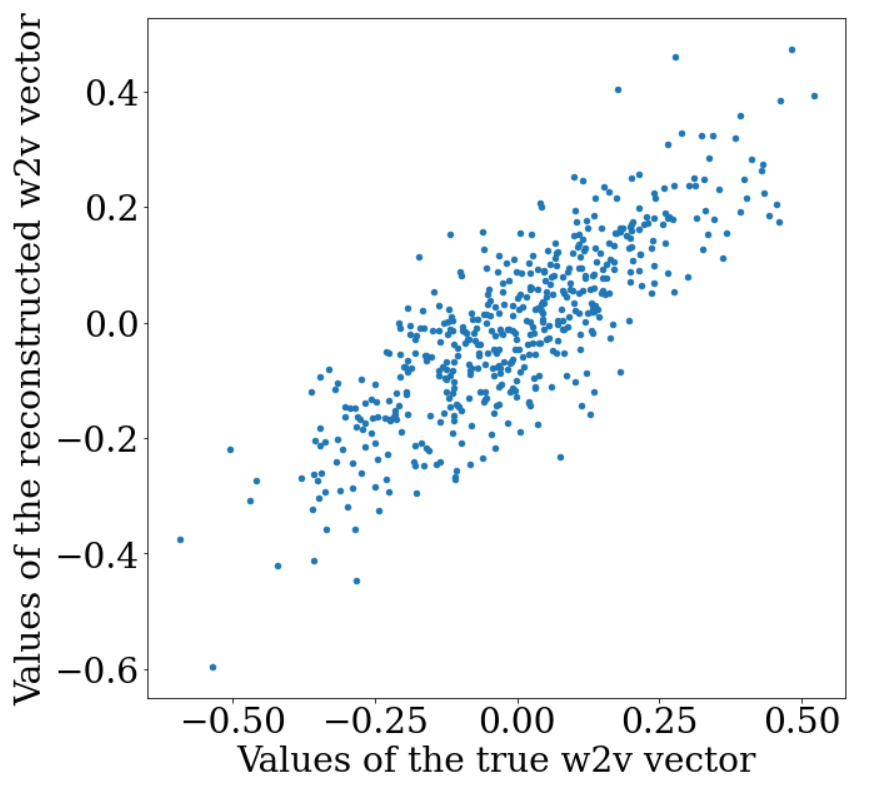}
\caption{Figure 1b: Plot of the values of the word2vec embedding for \emph{king}, versus  coefficients for the row of $C^\dag\cdot$ PMI corresponding to \emph{king}, for word2vec matrices trained on the same corpus (\emph{text8}). The Pearson correlation is one of the best possible at 0.825.} 
\label{plot link}
\end{subfigure}
\end{figure}

 %
To assess this, we use the \emph{text8} corpus \footnote{
A text dataset composed of 100 million characters from Wikipedia: \cite{mahoney_2006}.} both to train word2vec embeddings \footnote{Skip-gram architecture with negative sampling (1 word), negative sampling exponent equal to 1, no undersampling of common words, and a high dimension size of 500. These parameters allow us to be as close as possible to a direct factorization of the PMI matrix.} and to estimate a PMI matrix. We replace infinite values in the PMI matrix by 0. 
In Figure \ref{plot link spearman}, we show the distribution of the Pearson correlation coefficient (assessing the presence of a linear relation) between the word2vec embedding and the corresponding row of $C^\dag\cdot$ PMI  for the top ten thousand words in the corpus. As can be seen from the figure, the correlation tends to be  between 0.5 and 0.8. 
For instance in Figure \ref{plot link}, the word2vec embedding for \emph{king} is plotted against the row of $C^\dag\cdot$ PMI corresponding to \emph{king}. 


While the relation is not perfectly linear---many words have a correlation of around 0.55, far lower than that of \emph{king}---the empirical relations shown here leave open the possibility that it may indeed be ``sufficiently linear'' to be taken for granted.
However, while linearity is necessary for the result of \citet{allen} to go through, it is not sufficient. In the next section, we assess the critical question of whether the paraphrase error is small enough to serve as an explanation for the success of linguistic analogies.

\section{Empirical analysis of the error terms}

We now seek to examine the proposed explanation by calculating the proposed error terms empirically. However, in practice, many of the terms are undefined, since they rely on  cooccurrences unattested in practical corpora. 
The most extreme situation occurs when the two words of a paraphrase $\mathcal{W}=\{w_1,w_2\}$ are \emph{never} present in the same context window in the corpus.  We found that only 16\% of the paraphrase sets associated with the BATS analogy set \cite{GladkovaDrozd2016}---for example, \emph{king, woman}---were present together in the \emph{text8} corpus in a context window of length five. 
We refer to such paraphrase sets as ``well-defined'' with respect to the corpus. The problem of zero co-occurrence counts was anticipated by \citet{allen}, who propose to restrict their analysis to the case where the context window is sufficiently large that all relevant terms are well defined. We stress that our trained word2vec vectors are also trained with a context window of five, and yield expected levels of performance on the BATS analogy test, despite having access to  little training data on which to model co-occurrences such as \emph{king, woman}, \emph{queen, man}, and so on.

\begin{table*}[ht]
\centering
\scalebox{.64}{
\begin{tabular}{p{4.7cm}||p{.9cm}|p{.9cm}|p{.9cm}|p{.9cm}|p{.9cm}|p{.9cm}|p{.9cm}|p{.9cm}|p{.9cm}|p{.9cm}|p{.9cm}|p{.9cm}|p{.9cm}|p{.9cm}|p{.9cm}}
\hline \textbf{Category} 
& \textbf{I01}&\textbf{I02}&\textbf{I05}&\textbf{I06}&\textbf{I07}&\textbf{I08}&\textbf{I09}&\textbf{I10} & \textbf{D02}&\textbf{D03}&\textbf{D05}& \textbf{D08}&\textbf{D10}& \textbf{E01}& \textbf{E02}
\\ \hline 
Paraphrase error norm   &  177 & 153 & 111 & 127 & 126 & 124 &138 &97   &102 & 122 & 130 &110 & 107 & 124 &  176  \\ 
Dependence errors sum norm& 1006 & 938 & 867 & 903 & 957 & 883 &952 &908& 856 & 893 & 514 & 585 &699 & 749  & 848 \\ 
All errors sum norm  &  1032 &957 & 878 & 917 & 970 & 897 & 966 & 916 & 864 &905 & 539 & 602 & 710  &   765 & 875 \\ 
\hline 
\end{tabular}}
\centering

\centering

\scalebox{.64}{
\begin{tabular}{p{4.7cm}||p{.9cm}|p{.9cm}|p{.9cm}|p{.9cm}|p{.9cm}|p{.9cm}|p{.9cm}|p{.9cm}|p{.9cm}|p{.9cm}|p{.9cm}|p{.9cm}|p{.9cm}|p{.9cm}|p{.9cm}}
\hline \textbf{Category} &
\textbf{E03}&\textbf{E04}&\textbf{E05}& \textbf{E08}&\textbf{E09}&\textbf{E10}& \textbf{L02}&\textbf{L03}&\textbf{L04}&\textbf{L05}&\textbf{L06}&\textbf{L07}& \textbf{L08}&\textbf{L09}&\textbf{L10}
\\ \hline 
Paraphrase error norm   & 162 &176 & 155 & 229 & 179 & 190  &197 & 189 & 209 & 206 & 133 & 169 & 185 &175 & 432   	 \\ 
Dependence errors sum norm &866 & 797 & 519 & 739 &910 & 833 & 642 & 982 &907 & 1103 & 921 &995 & 1044 & 1017 &1302  \\ 
All errors sum norm  &889 & 822 &553 & 778 & 933 & 865 &  683 &1007 &939 & 1131 & 937 &1016 & 1066 & 1040 & 1416  \\ 
\hline 
\end{tabular}}
\caption{L2 norms of the error terms in \ref{eq_allen}, following our implementation.}
\label{tab:paraphrase tests norm}
\end{table*}

\begin{table*}[ht]
\centering
\scalebox{.625}{
\begin{tabular}{p{2.2cm}||p{1.1cm}|p{1.1cm}|p{1.1cm}|p{1.1cm}|p{1.1cm}|p{1.1cm}|p{1.1cm}|p{1.1cm}|p{1.1cm}|p{1.1cm}|p{1.1cm}|p{1.1cm}|p{1.1cm}|p{1.1cm}|p{1.1cm}}
\hline \textbf{Category} 
& \textbf{I01}&
\textbf{I02}&
\textbf{I05}&
\textbf{I06}&
\textbf{I07}& \textbf{I08}&
\textbf{I09}&
\textbf{I10} & 
\textbf{D02}&
\textbf{D03}&
\textbf{D05}&
 \textbf{D08}&
\textbf{D10}& 
\textbf{E01}& \textbf{E02}
\\ \hline 
Average rank & 
7762K&
7589K&
7759K&
8744K&
8160K&
6454K&
7028K&
11889K&
31952K&
19558K&
7857K&
1506K&
2556K&
4394K&
9507K\\ 

Median rank & 
1630K&
2195K&
3055K&
3239K&
2530K&
4090K&
3004K&
4535K&
6754K&
3564K&
3260K&
1506K&
2556K&
2117K&
1622K\\ \hline 
\hline 
\end{tabular}}
\centering

\centering

\scalebox{.625}{
\begin{tabular}{p{2.2cm}||p{1.1cm}|p{1.1cm}|p{1.1cm}|p{1.1cm}|p{1.1cm}|p{1.1cm}|p{1.1cm}|p{1.1cm}|p{1.1cm}|p{1.1cm}|p{1.1cm}|p{1.1cm}|p{1.1cm}|p{1.1cm}|p{1.1cm}}
\hline \textbf{Category} &
\textbf{E03}&
\textbf{E04}&
\textbf{E05}& \textbf{E08}&
\textbf{E09}&
\textbf{E10}& \textbf{L02}&
\textbf{L03}&
\textbf{L04}&
\textbf{L05}&
\textbf{L06}&
\textbf{L07}& \textbf{L08}&
\textbf{L09}&
\textbf{L10}
\\ \hline 
Average rank & 
1305K&
5611K&
9192K&
727K&
8421K&
11946K&
52183K&
1857K&
12687K&
6747K&
2475K&
7727K&
4502K&
4679K&
16871K\\ 

Median rank & 
695K&
1703K&
1426K&
854K&
1908K&
169K&
52182K&
1261K&
2460K&
1343K&
2255K&
2136K&
1549K&
1 739K&
785K\\ \hline 

\hline 
\end{tabular}}


\caption{For an analogy equivalent to two paraphrases $\mathcal{W}$ and $\mathcal{W_*}$, the rank of $\mathcal{W_*}$ in the list of the closest paraphrases to $\mathcal{W}$ with respect to the L2 norm of the paraphrase error vector. 7762K means a rank of 7762000, rounded to the nearest thousand.}
\label{tab:paraphrase tests rank}
\end{table*}


At a minimum, if the proposed explanation holds, the cases for which the error terms are empirically well-defined should show signs of the paraphrase error being relatively small. We now detail how we implemented the error terms in cases for which they were well-defined. We count co-occurrences $N(w_i,w_j,w_k)$ in \emph{text8} for all triplets of words $w_i,w_j,w_k$, with $w_k$ at the center of the context window, and $W=\{w_i,w_j\}$ any paraphrase, both occurring anywhere within a context window of width five. We restrict analysis to the ten thousand most frequent word types $w_i$ and $w_j$, yielding $10^8$ possible paraphrases.\footnote{$w_k$ is allowed to vary over all of the types included in the training for word2vec, of which there are 71290. Thus, for each paraphrase, the error vectors have 71290 elements, one for each vocabulary word.} 
We use the relative frequencies as estimators of $p(w_k|\{w_i,w_j\})$ and $p(\{w_i,w_j\}|w_k)$, and  marginalize to obtain $p(w_i|w_k)$, $p(\{w_i,w_j\})$ and $p(w_k)$. The error terms follow.
Since this can still lead to  ill-defined elements, we replace  $log(+\infty)$ and $log(0)$ by $+/- log(\epsilon)$, with $\epsilon=10^{-15}$ (within reason, the value of $\epsilon$ is immaterial). We also replace $log(0/0)$ with 0. 

Table \ref{tab:paraphrase tests norm} shows  the mean  and median values 
of the L2 norms of the paraphrase error vectors across several  categories of the BATS dataset. We compare them with the sum of the four dependence error terms (the dependence error reflects statistical dependencies within $\mathcal{W}$ and $\mathcal{W_*}$ irrelevant to the analogy), as well as the sum of all five error terms (equal to the difference between the PMI of $\mathcal{W}$ and  $\mathcal{W_*}$).
The paraphrase error is indeed smaller than the other error terms. 
However, as we  now show, the paraphrase error is not small \emph{enough} to contribute substantially to the success of  analogies.\footnote{We note also that the error values seem relatively consistent between categories, while success on the analogy test varies differ greatly between categories.}

Take the  norm of the paraphrase error vector $\rho$ as a measure of the divergence in the PMI between two paraphrases. For an analogy with associated paraphrases $\mathcal{W}$ and $\mathcal{W_*}$, we assess how many paraphrases are closer to $\mathcal{W}$ than to $\mathcal{W_*}$ by calculating the rank of the norm of $\rho^{\mathcal{W,W_*}}$ among all $\rho^{\mathcal{W,X}}$, where $\mathcal{X}$ spans over all pairs of words constructible from the top ten thousand most frequent words in the corpus. 
To do so, we define a Paraphrase Conditional Information matrix (PCI). 
For $W_{ij}=\{w_i,w_j\}$ and $w_k$,  we define  $PCI(l_{ij}, k)$, the value at column $l_{ij}$ and row $k$  to be $log(p(W_{ij}|w_k))$, where
with $l_{ij}$ is a unique index associate with tuple $(i,j)$. We compute only the positive PCI, to obtain a sparse matrix. 
The difference between two PCI columns is  a paraphrase error vector, and their Euclidean distance is the norm of the paraphrase error.


We now compute, for each analogy, the distance between the PCI column of $\mathcal{W}$ and every other column (paraphrase) of the PCI matrix. 
We calculate  the rank of the true analogy pair $\mathcal{W_*}$.  Given that the analogy test generally succeeds in picking out $b$ as being the most similar to $a-a^*+b^*$ out of the entire vocabulary (modulo  \citealt{Linzen_2016_Issues_in_evaluating_semantic_spaces_using_word_analogies}), we would expect that, for successful analogies, the paraphrase error for the true analogy would be among the highest, if small paraphrase error were the explanation for success. Table \ref{tab:paraphrase tests rank} displays the mean 
of this rank within each BATS category. The rank is extremely low (in the millions), making the paraphrase error in true analogies  far too high to be the explanation for their success.\footnote{Limiting the search to the paraphrases composed by at least one of the words of $W_*$ still results in a very low rank for $W_*$.}

\section{Conclusion}


 
Recent work has  shown that, in spite of the standard analogy test's confound with  simple vector similarity, distributional word vectors genuinely do encode linguistic regularities as directional regularities above and beyond vector similarity  \citep{fournier-etal-2020-analogies}. Further research is warranted into the mechanisms by which distributional word embeddings come to show  these regularities.
However, the analysis of analogies as paraphrases does not hold up as an explanation of 
performance on the analogy test---nor would an explanation of performance on the  \textsc{3CosAdd}  analogy test be a satisfying result, since the test is not a useful measure  to begin with.



\section*{Acknowledgments}

This work was funded in part by the European Research Council (ERC-2011-AdG-295810 BOOTPHON), the Agence Nationale pour la Recherche (ANR-17-EURE-0017 Frontcog, ANR-17-CE28-0009 GEOMPHON, ANR-10-IDEX-0001-02 PSL*, ANR-19-P3IA-0001 PRAIRIE 3IA Institute, ANR-18-IDEX-0001 U de Paris, ANR-10-LABX-0083 EFL) and grants from CIFAR (Learning in Machines and Brains), Facebook AI Research (Research Grant), Google (Faculty Research Award), Microsoft Research (Azure Credits and Grant), and Amazon Web Service (AWS Research Credits).

\bibliography{eacl2021} 
\bibliographystyle{acl_natbib}

\end{document}